\documentclass{article}

\usepackage{preprint}

\usepackage[utf8]{inputenc} 
\usepackage[T1]{fontenc}    
\usepackage{hyperref}       
\usepackage{url}            
\usepackage{booktabs}       
\usepackage{amsfonts}       
\usepackage{nicefrac}       
\usepackage{microtype}      
\usepackage{lipsum}

\usepackage{booktabs} 
\usepackage{subfig}
\usepackage{amsmath}
\usepackage{mathtools}
\usepackage[toc,page]{appendix}
\usepackage{algcompatible}
\usepackage{algorithm}
\usepackage{varwidth}
\usepackage{longtable}
\usepackage{comment}
\usepackage{enumerate}
\usepackage[toc,page]{appendix}
\algnewcommand\algorithmicforeach{\textbf{for each}}
\algdef{S}[FOR]{ForEach}[1]{\algorithmicforeach\ #1\ \algorithmicdo}
\usepackage{xcolor}
\usepackage[noend]{algpseudocode}
\DeclareCaptionType{mycapequ}[Equation][]
\algnewcommand{\LineComment}[1]{\State \(\triangleright\) {\footnotesize #1}}
\algnewcommand{\algorithmicand}{\textbf{ and }}
\algnewcommand{\algorithmicor}{\textbf{ or }}
\algnewcommand{\OR}{\algorithmicor}
\algnewcommand{\AND}{\algorithmicand}
\algnewcommand{\var}{\texttt}
\algnewcommand\algorithmicforeach{\textbf{for each}}
\algdef{S}[FOR]{ForEach}[1]{\algorithmicforeach\ #1\ \algorithmicdo}
\usepackage{hyperref}
\usepackage{soul}
\newcommand{\pluseq}{\mathrel{+}=}
\newcommand{\etal}{{\it et al}}

\newcommand{\ie}{{\textit{i.e.}}}

\newcommand{\examm}{EXAMM }

\definecolor{darkgreen}{rgb}{0,0.3922,0}

\newcommand{\citet}[1]{\citeauthor{#1} \shortcite{#1}} 
\newcommand{\citep}{\cite}

\newcommand{\customfootnotetext}[2]{{
  \renewcommand{\thefootnote}{#1}
  \footnotetext[0]{#2}}}

\title{The Ant Swarm Neuro-Evolution Procedure for Optimizing Recurrent Networks}

\author{
  AbdElRahman A. ElSaid\\
  \texttt{aelsaid@mail.rit.edu} \\
   \And
  Alexander G. Ororbia** \\
  \texttt{ago@cs.rit.edu} \\
  \\
    Golisano College of Computing and Information Sciences\\
    Rochester Institute of Technology\\
    Rochester, NY 14623\\
  \And
  Travis J. Desell**\\
  \texttt{tjdvse@rit.edu} \\
}

\begin{document}
\maketitle

\begin{abstract}
\label{sec:abstract}
Hand-crafting effective and efficient structures for recurrent neural networks (RNNs) is a difficult, expensive, and time-consuming process. To address this challenge, we propose a novel neuro-evolution algorithm based on ant colony optimization (ACO), called ant swarm neuro-evolution (ASNE), for directly optimizing RNN topologies.  The procedure selects from multiple modern recurrent cell types such as $\Delta$-RNN, GRU, LSTM, MGU and UGRNN cells, as well as recurrent connections which may span multiple layers and/or steps of time. In order to introduce an inductive bias that encourages the formation of sparser synaptic connectivity patterns, we investigate several variations of the core algorithm. We do so primarily by formulating different functions that drive the underlying pheromone simulation process (which mimic L1 and L2 regularization in standard machine learning) as well as by introducing ant agents with specialized roles (inspired by how real ant colonies operate), i.e., \emph{explorer ants} that construct the initial feed forward structure and \emph{social ants} which select nodes from the feed forward connections to subsequently craft recurrent memory structures. We also incorporate a Lamarckian strategy for weight initialization which reduces the number of backpropagation epochs required to locally train candidate RNNs, speeding up the neuro-evolution process. 
Our results demonstrate that the sparser RNNs evolved by ASNE significantly outperform traditional one and two layer architectures consisting of modern memory cells, as well as the well-known NEAT algorithm. Furthermore, we improve upon prior state-of-the-art results on the time series dataset utilized in our experiments.
\end{abstract}\customfootnotetext{**}{Indicates equal advising.}

\section{Introduction}
\label{sec:intro}
Given their success across a wide swath of pattern recognition tasks, artificial neural networks (ANNs) have become a popular tool to use when attempting to solve data-driven problems. However, in order to solve increasingly more complicated problems, neural architectures are becoming vastly more complex. Increasing the complexity of an ANN entails having to operate with more layers of neural processing elements required, most of which are usually wider and more densely-connected, greatly complicating the model design process. The resulting increase in complexity introduces new challenges and complications when fitting these ANN models to actual data. These problems are further compounded when ANNs are meant to process temporal data, entailing recurrent connections which can span varying periods of time. As a result, crafting performant ANNs becomes expensive and incredibly difficult for engineers, highlighting a grand challenge facing the domain of machine learning -- the automation of ANN architecture design, which includes selecting the form of the underlying synaptic topology as well as the values of the weights themselves. The key to this automation might lie in developing optimization procedures that can effectively explore the vast, combinatorial search space of possible topological structures that could be constructed from a large set of neuronal units and the wide variety of synaptic connectivity patterns that relate them to one another. 

Recent interest in automated architecture search has resulted in many proposed ideas related to deep feed forward and convolutional networks, including those based on nature-inspired metaheuristics \cite{yang2010nature}. However, few, if any, have focused on the far more difficult problem of optimizing recurrent neural networks (RNNs) aimed at processing temporal, sequential data such as time series, i.e., automated RNN design.

This study addresses the challenge of automated RNN design by developing a novel ANN topology optimizer based on concepts from artificial evolution and ant colony optimization (ACO). Specifically, we propose an algorithm called Ant Swarm Neuro-Evolution (ASNE), which automatically constructs and optimizes the topology of RNNs, with a focus on time series data prediction. In developing our optimization approach, we furthermore develop and experiment with variations of our method in the following ways:
\begin{itemize}
    \item In order to encourage the discovery of more sparsely-connected neural topologies, we investigate different schemes for dynamically modifying the pheromone traces deposited by ant agents that compose the swarm. Specifically, we introduce functions for introducing regularization into the overall optimization, slowly clearing out densely-connected synaptic areas by depriving poorly performing weights/edges of pheromone accumulation.
    \item We incorporate and analyze various weight initialization schemes and find that a Lamarckian inheritance strategy is highly effective.
    \item Inspired by the role-specialization that ants operate under within the context of real-world ant colonies, we extend ASNE to utilize different specialized ant agents to modularize the underlying synaptic connectivity construction process, which we find greatly improves solutions found by our metaheuristic.
\end{itemize}
Experimentally, we validate our proposed nature-inspired metaheuristic on an open-access real-world time series data set collected form a coal-fired power plant. A rigorous ablation study of the ASNE algorithm 
is conducted by analyzing the candidate network topologies it finds. A total over $1600$ experiments with varying heuristics and hyperparameters were performed, which entailed training $32,000,000$ different RNNs. Our results indicate that ASNE is able to build well performing, arbitrary RNN structures with connections that span both structure and time using both simple and complex memory cells. More importantly, ASNE is shown to significantly outperform the well-known neuro-evolutionary algorithm, NEAT~\cite{stanley2002evolving}, as well as the state-of-the-art evolutionary optimizer, EXAMM~\cite{ororbia2019examm}, which have held the prior best results on this data set.
\section{Related Work}
\label{sec:related_work}
With respect to neuroevolution of recurrent network topologies, a great deal of work already exists, ranging from stochastic alteration of the topology as in drop-out~\cite{srivastava2014dropout} to something more sophisticated like that in the original NEAT~\cite{stanley2002evolving} and its more modern incarnate HyperNEAT~\cite{stanley2009hypercube}. Other proposed approaches include EPNet~\cite{yao1997new}, EANT~\cite{kassahun2005efficient}, GeNet\cite{xie2017genetic}, CoDeepNEAT~\cite{miikkulainen2019evolving}, and EXACT~\cite{desell2017large}. EXACT was recently extended to evolve RNNs that used LSTM memory cells (named EXALT) and shown to perform quite well on time-series prediction problems~\cite{elsaid2019evolving}. Later, the algorithm, named EXAMM, was generalized to evolve networks consisting of a library of recurrent memory cells~\cite{ororbia2019examm}. These previously proposed ideas center around the use of a genetic algorithm \cite{holland1992adaptation}, where optimization is inspired by approaches that draw from the evolution of organisms, of either Darwinian and/or Lamarckian nature. More recently, work by Camero \etal have shown that a Mean Absolute Error (MAE) random sampling strategy can provide good estimates the performance of RNNs~\cite{camero2018low} and have successfully used it instead of actually evaluating or training RNNs to speed up neuro-evolution of LSTM RNNs~\cite{camero2019specialized}. 

Nonetheless, very few studies in the body of work described above consider ant colony optimization (ACO) \cite{dorigo1992optimization} as the central optimizer for network topology, and even fewer in general focus on exploring how to evolve complex temporal models like the RNN, with a few exceptions, such as EXALT and EXAMM. Of the few that have investigated ACO, most existing work has used it to strictly optimize feed forward networks and, even in that case, have dominantly focused on either initializing the weights of the connections~\cite{mavrovouniotis2013evolving}, or on reducing the dimension of the input vector solution space~\cite{sivagaminathan2007hybrid}. One notable effort that has used ACO for RNN optimization in some form is \cite{desell2015evolving}, which used ACO to optimize smaller neural network structures based on Elman recurrent networks \cite{elman1990finding}.  

This paper contributes to the domain of nature-inspired neural network topology optimization by proposing a novel metaheuristic for evolving the full structure of an RNN as opposed to prior studies that have applied the technique as only a partial component of the optimization process~\cite{elsaid2018optimizing} or in smaller Elman RNN topologies with limited recurrent connectivity~\cite{desell2015evolving}. Furthermore, our algorithm is capable of utilizing the same full suite of recurrent memory cells as the state-of-the-art evolutionary algorithm EXAMM (LSTM, GRU, MGU, UGRNN, and $\Delta$-RNN cells). To the best of our knowledge, we are the first to propose an ACO-based approach to automate RNN design, offering a powerful procedure that combines concepts of both neuro-evolution and ant colony metaheuristic optimization.
\section{Ant Swarm Neuro-Evolution (ASNE)}	
\label{sec:method}
ASNE handles the optimization of ANN structures by constructing a simple multi-agent system, where each agent treats the ANN as graph structure, considering neuronal processing elements (PEs) as the nodes and the synaptic weights that connect PEs as the edges. In order to design the operations that these agents perform as well as the manner in which they traverse the ANN graph, we may appeal to the metaphor of ants and the collective they holistically form, \ie, the ant colony. As a result, the agents will function based on simplifications of myrmecological principles, such as the mechanics of ant-to-ant social interaction.

At a high level, in ASNE, the individual ant agents operate on a single massively connected ``superstructure'', which contains all possible ways that PEs may connect with each other both in terms of structure, \ie, all possible feedforward pathways that start from the input/sensory PEs and end at the output/actuator PEs, and time, \ie, all possible recurrent connections that span many multiple time delays. In our implementation, ants choose to move over connections between nodes (or neurons), probabilistically and as a function of a simulated chemical known as the ``pheromone''. In nature, the pheromone is one primary driver of how ants communicate with each other, the traces of which allow the collective to ``know'' of potential food sources ensuring the survival of the colony in the long term. When an ant finds food, the ant will start marking the path it takes to return back to the colony, the pheromone trace of which other ants will then subsequently follow. In the ANN superstructure, these traces, which are simulated by an additional, dynamic scalar weight (or importance value) assigned to a given synapse, will bias any given ant agent to favor selecting some possible (more rewarding) synaptic pathways over others. 


\begin{algorithm}
	\footnotesize{}
    \caption{Ant Colony Algorithm}\label{antcol_absract_pseudo}
    \begin{algorithmic}
		\Procedure{Master}{}
			\LineComment{construct fully connected structure with edges holding initial pheromone and weights values.}
			\State $colony = \textbf{new}~Colony$
			\For {$i \gets 1 \dots max\_iteration$}
				\State $nn_{new} \gets ants\_swarm( colony )$
				\State $send\_to\_worker (nn_{new}, worker.id)$
				\State $nn_{new}, fit \gets receive\_fit\_from\_worker( )$
				\If { $nn\_fitness< worst\_population\_member$ }
					\State $population.pop ( worst\_population\_member)$
					\State $population.add ( new\_nn )$
					\State $reward\_paths\_in\_colony ( new\_nn )$
				\EndIf			
				\If {$use\_Lamarckian\_weight\_inheritance$}
							\If {$use\_phi\_function$}
								\LineComment {update colony's weights from $nn_{new}$ using phi equation}
							\ElsIf {$constant\_phi$}
								\LineComment {update colony's weights with constant fraction of $nn_{new}$ weights}
							\EndIf
				\EndIf
				\If {$bias\_forward\_paths$}
					\LineComment{Sum (Fwd Edge/Recurrent Pheromone) = Sum(Bkwd Recurrent Edges Pheromone)}
				\EndIf				$periodic\_pheromone\_evaporation ( )$
			\EndFor
		\EndProcedure
		
		\Procedure{Worker}{}
			\State $recieve\_from\_master (nn)$		 
			\State $fitness \gets train\_test\_nn ( nn )$
			\State $send\_fitness\_to\_master ( nn, fitness )$
		\EndProcedure

		\Procedure {$Ants\_Swarm$} {}
			\If {$one\_layer\_jump$}
				\LineComment {ants only move one layer at a step}
			\ElsIf {$\quad ! one\_layer\_jump$}
				\LineComment {ants can jump over layers}
			\EndIf
			\If {$one\_ant\_species$}
				\For {$ant \gets 1 \dots no\_ants$}
					\LineComment {ant chooses the nodes, edges, and recurrent edges 
					}
				\EndFor
			
			\ElsIf {$two\_ants\_species$}
				\For {$ant \gets 1 \dots no\_ants/2$}
						\LineComment{ant choose the nodes, edges from colony}
				\EndFor
				\If {$social\_forward \And ! social\_backward$}
					\For {$ant \gets 1 \dots no\_ants/2$}
						\LineComment{ants choose rec\_edges only from fwd rec\_edges}
					\EndFor
				\ElsIf {$social\_backward  \And ! social\_forward$}
					\For {$ant \gets 1 \dots no\_ants/2$}
						\LineComment{ants choose rec\_edges only from bwd rec\_edges}
					\EndFor
				\ElsIf {$social\_forward \And social\_backward$}
					\For {$ant \gets 1 \dots no\_ants/4$}
						\LineComment {ants choose rec\_edges only from fwd rec\_edges}
					\EndFor
					\For {$ant \gets 1 \dots no\_ants/4$}
						\LineComment {ants choose rec\_edges only from bwd rec\_edges}
					\EndFor
				\EndIf
			\EndIf
			
			\Return new\_nn
		\EndProcedure

		
		
		\Procedure {$Reward\_Paths$} {$nn$} 
			\ForEach {$edge \in nn.edges$} 
				\If {$use\_constant\_reward$}
					\State {$colony.edge[ edge.id ].pheromone \pluseq constant $}				
				\ElsIf {$use\_fitness$}
					\State {$colony.edge[ edge.id ].pheromone  = \text{Eqn~\ref{eq:reward_fit}}$}
				\ElsIf {$use\_L1\_regularization$} 
					\State {$colony.edge[ edge.id ].pheromone  = \text{Eqn~\ref{eq:reward_fit_weight_l1}}$}
				\ElsIf {$use\_L2\_regularization$}
					\State {$colony.edge[ edge.id ].pheromone = \text{Eqn~\ref{eq:reward_fit_weight_l2}}$}
			\EndIf
		\EndFor
	\EndProcedure
	
\end{algorithmic}
\end{algorithm}

The few existing efforts on using forms of ACO for RNN optimization \cite{elsaid2018optimizing,elsaid2019evolving} restrict the ACO process to operate within individual LSTM memory cells. In contrast, ASNE allows individual ants traverse a single massively connected ``superstructure'', which contains all possible ways that the nodes of an RNN may connect with each other both in terms of structure (\ie, all possible feed forward connections), and in time (\ie, all possible recurrent connections spanning many multiple time delays)\footnote{Note that this superstructure is more connected than a standard fully connected neural network -- each layer is also fully connected to each other layer as well, allowing for forward and backward layer skipping connections, with additional recurrent connections between node pairs for each time skip allowed.}. The high-level pseudocode for our ASNE topology optimizer is depicted in Algorithm~\ref{antcol_absract_pseudo}.

ASNE was developed as an asynchronous parallel system for use on high performance computing resources, which has a master process that maintains the colony information and worker processes to (locally) train the RNNs. This parallel implementation is asynchronous, the master process generates new RNNs as needed for worker processes (which operate on separate, dedicated CPU or GPU resources) and updates colony information and pheromones as trained RNN results are returned.  This results in a naturally load balanced algorithm with high scalability.

Within the master process itself, ASNE operates by having a fixed number of ant agents traverse the neural superstructure. Ants choose to move over connections between nodes (neurons) randomly, but they are probabilistically biased towards connections with higher simulated ``pheromone'' values. Pheromone deposit values are periodically evaporated to prevent the search process from becoming stuck in local minima.
Interestingly enough, the modification of the evaporation function could be considered to a way in which one could encode certain priors into the ANN itself.

From the overall superstructure, which the ant agents exclusively operate on, RNN subnetworks are extracted (as dictated by the current pheromone trace network available at the current simulation time step, which yields a map of nodes and connecting synapses, both recurrent and feedforward, visited by the ant agents) and then further fine-tuned locally with only a few epochs of back-propagation (backprop) through time. After a particular worker is done locally training a RNN subnetwork, the candidate's weight values and cost (fitness) function (measure on a validation subset of data) are communicated back to the swarm and superstructure (housed in the master process), adjusting the pheromone trace network and affecting future ant agent traversal behavior. 

One crucial element in our ASNE procedure is the introduction of different ant agent types, which is inspired by how real ants specialize to act according to specific roles to serve the needs of the colony \cite{odonnell2018antroles}.
Specifically, we consider designing ant agents that serve specific roles in constructing parts of candidate RNN subnetworks -- some ants exclusively traverse feedforward synaptic pathways while others only explore recurrent synaptic pathways. 
The high-level pseudocode for our ASNE topology optimizer is depicted in Algorithm \ref{antcol_absract_pseudo}\footnote{The full code is posted at: \url{https://github.com/travisdesell/exact/tree/adding_ant_colony}.}.

Within the framework of ASNE, we investigate variations of its various underlying mechanisms. These include the use of Lamarckian weight initialization, allowing ant agents to also select from multiple memory cell types as opposed to operating exclusively with simple neurons, introducing specialized ants that have different graph traversal strategies, and constraining ant movement and manipulating the pheromone evaporation function in order to encourage the discovery of sparse RNN topologies.
    
\subsection{Lamarckian Weight Initialization}
\label{sec:lamarck}
Edges and recurrent edges' weights can be randomly initialized each time a new RNN is generated by the ants. However, initializing parameters this way requires local tuning (via backprop) for many epochs for the RNN to reach suitable generalization error, as they do not make use of any information gained by prior trained RNN candidates. Further, it has been shown that reusing of parental weights (\ie, epigenetic or Lamarckian weight initialization) can significantly speed up the neuro-evolution process and result in better performing, smaller ANNs in general~\cite{desell2018accelerating}.

To apply Lamarckian weight initialization to ASNE, each edge in the ant swarm's connectivity super-structure also tracks a weight value in addition to its pheromone value.  These weights are randomly initialized uniformly $U(-0.5,0.5)$. Each time a generated RNN performs well, the weights of its best performance, as measured on a validation data subset, are used to update the weight values in the swarm's super-structure internal bookkeeping.

Formally, we define $\Phi$ is a function of the population's best and worse evaluated RNN fitness, $W_{{colony}_i}$ as the colony's edge weight,, $W_{{RNN}_i}$ as the corresponding neural network's edge weight,  $fit_{pop\_best}$ as the population's best fitness, and $fit_{pop\_worst}$ as the population's worst fitness. Weight initialization then proceeds as follows:
\begin{subequations}
\begin{align}
x &= \frac{fit_{new} - fit_{pop\_best}}{fit_{pop\_worst} - fit_{pop\_best}}\\
\Phi(fit_{new})    &= min \Bigg( max \Big((1-x), 0\Big), 1\Bigg)\\
W_{{colony}_i}     &= \Phi W_{{RNN}_i} + (1-\Phi) W_{{colony}_i} \mbox{.}
\end{align}
\label{eq:function_phi}
\end{subequations}
\noindent
With respect to the function $\Phi$, we investigated two variations.
The first variant, as shown in Equation~\ref{eq:function_phi}, used the fitness of the RNN used to update the weights to determine how much these new (locally found) weight values effect those of the colony. The second variant of $\Phi$ was set to a predetermined constant instead of being calculated or adjusted by fitness. This process essentially allows for a running average (either with a fixed update or dynamic update based on fitness) of the best weights found for each connection in the superstructure.  
When a new RNN is generated, it uses the current weight values in whatever edges that were extracted from the superstructure on the master process. This allows for a Lamarckian evolution for edges weights, as prior RNNs with the best fitness scores are allowed to pass on their weights to future generations. 

\subsection{Memory Cell Selection}
\label{sec:memcell_select}
For any particular node in the super-structure, ASNE also has the ability to utilize the pheromones present to select which memory cell type a particular node will be in the generated network. A node could chosen to be either an LSTM~\cite{hochreiter1997long}, a GRU~\cite{chung2014empirical}, an MGU~\cite{zhou2016minimal}, a UGRNN~\cite{collins2016capacity}, or a $\Delta$-RNN cell~\cite{ororbia2017learning}. We refer the reader to these works for the formulations of these memory cells. Pheromones are deposited and updated for each of these memory cell possibilities as described below.

\subsection{Altering Graph Traversal with Ant Species}
\label{sec:aco_graph_traversal}
As mentioned above, we explored various strategies for guiding ant traversal over the connectivity superstructure. Inspired by role specialization in real colonies, we implemented ant agents that explored the connectivity graph in specific ways. First, we started with a generic ant agent, called the \emph{standard ant}, which was allowed to traverse through the massively connected colony superstructure in an unbiased manner. This, in essence, recovers the standard simple ant agent in classic ACO, which has complete freedom to explore any piece of a given graph structure. 
However, it became quickly apparent that this type of ant would get ``stuck'' in the network, generating a significantly high number of recurrent connections before finally reaching an output node. This meant that the RNN candidates extracted for local fine-tuning were rather dense, and in turn, compute-heavy (featuring many extraneous parameters as is characteristic of over-parameterized models). 

\begin{figure}
\centering
\includegraphics[width=.4\textwidth,height=.1\textheight]{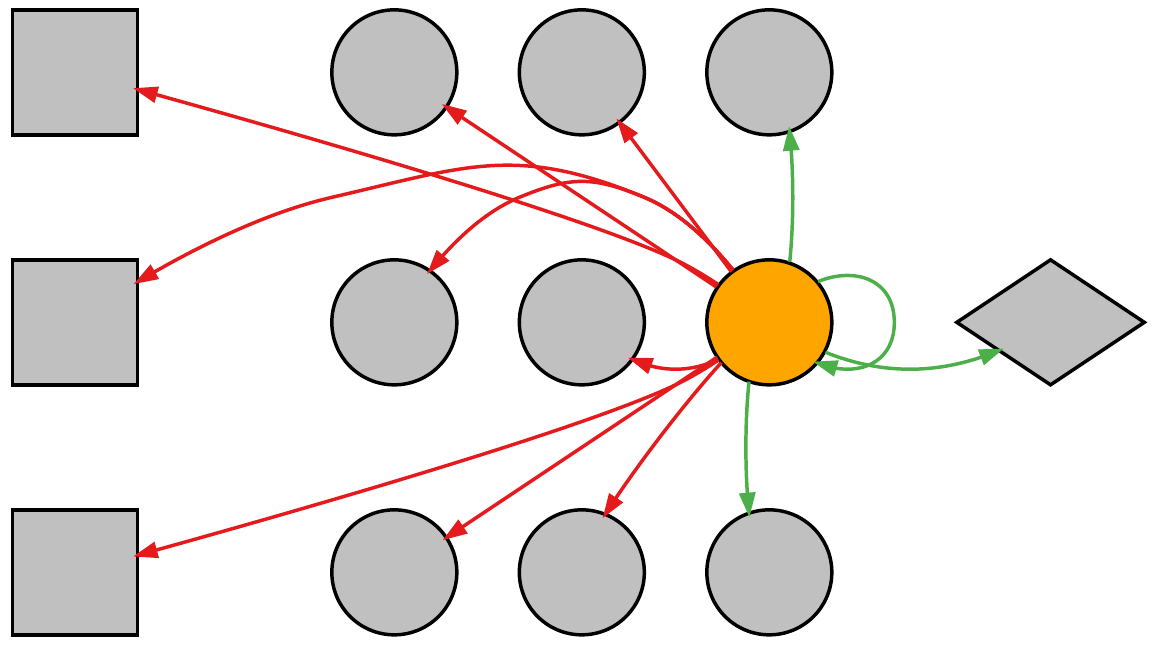}
\caption{\label{fig:more_bwd} Potential paths an ant can take from a given node (in orange) with the massively-connected superstructure. The number of recurrent paths (red) far outnumber the forward paths (green). This problem is exacerbated as the possible recurrent time scale increases, which results in multiple backward recurrent connections for each red connection, each going back a different number of time steps in the past.}
\end{figure}
	    
Why do standard/simple ants get stuck or meander too long in the superstructure?
In the superstructure, nodes (especially at the final hidden layer) have the option of selecting potential backward recurrent paths, which significantly outnumber the number of potential forward moving paths (see Figure~\ref{fig:more_bwd}). Assuming that each connection has an equal number of pheromones (which is a standard setting for pheromone initialization), agents will circle around the colony using these backward paths, yielding RNN candidates with very dense recurrent structure.

To prevent this problem, our first tactic was to alter the pheromone deposit function by adding extra pheromones to forward paths upon initialization as well as after every pheromone update. The biasing method yielded better proportions of forward and backward paths. Algorithm~\ref{forwad_bias_code} illustrates this process.

    \begin{algorithm}
    \footnotesize
    \caption{Forward Connections Bias Algorithm}\label{forwad_bias_code}
    \begin{algorithmic}
    \ForEach {$node \in Nodes$}
        \If {$fwd\_pheromone_{total} < 0.75 \cdot bwd\_pheromone_{total}$ \OR  $bwd\_edges > fwd\_edges$}
            \ForEach {$fwd\_edge \in Fwd\_Edges$}
            	\State $\tau_{fwd\_edge}	 \gets 
            	(\tau_{fwd\_edge}/{\tau_{fwd\_edges}^{total}}) \cdot \tau_{bwd\_edges}^{total} $
            \EndFor
        \EndIf
    \EndFor
    \end{algorithmic}
    \end{algorithm}

Even with this forward path bias added to the pheromone deposit function, when using standard ants, we found that ASNE still tended to favor the generation of fairly dense networks. Altering the number of ant agents used to explore the structure as a means to control density of RNN candidates proved to help somewhat but was rather unwieldy and entailed far too much external human intervention. 
Instead, we developed an ant agent role specialization scheme that we found worked far better as an automatic control mechanism to control the network size and synaptic density.

\begin{figure}
\centering
\includegraphics[width=.48\textwidth,height=.15\textheight]{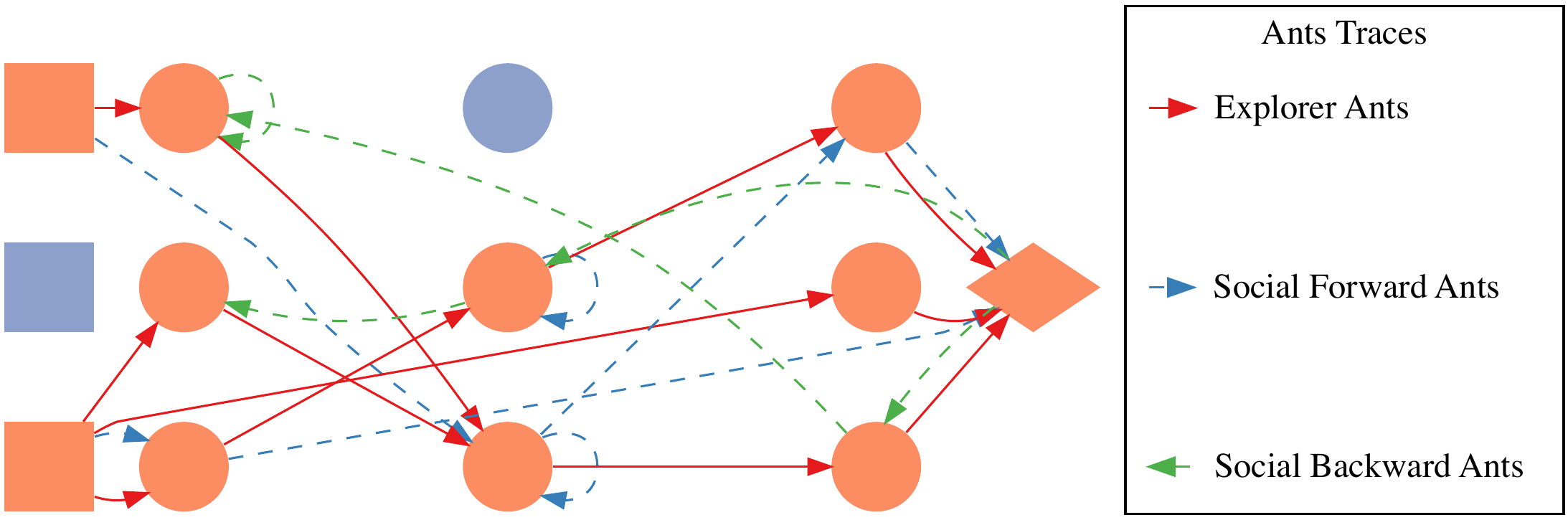}
\caption{\label{fig:ants_traces} In multi-role traversal, explorer ants (red) first select the forward paths in the network, creating a basic structure for the RNN. The social ant agents then select from the nodes chosen by the explorer ants. Within the social ant agent role, there is a sub-specialization consisting of forward recurrent ants (blue) that create additional forward recurrent connections between these nodes and backward recurrent ants (green) that move backwards from the output toward the input, creating backward recurrent connections between the same nodes.
}
\end{figure}

The first agent role, the \emph{explorer ant}, means that the agent is only allowed to choose from forward connections in the connectivity superstructure. The connections selected by this specialized agent would utilized to generate the base neural structure upon which recurrent connections could then be added to. 
After the explorer ants selected the possible nodes and forward connections, two additional specializations of what we call \emph{social ants} would then be used, {\it i)} \emph{forward recurrent ants} and, {\it ii)} \emph{backward recurrent ants}. Social ants are first restricted to only visiting nodes that have already been selected by the explorer ants. In the case of the forward recurrent ants, when a path is chosen, the agent would specifically create a recurrent connection that moved forward in the network along the same path, along with a selected time skip (determined by pheromones). Backward recurrent ants, on the other hand, move backwards through the network and, for each path they take, a backward recurrent connection is added, along with a selected time skip (also determined by pheromones). Figure~\ref{fig:ants_traces} provides an example of possible pathways that these specialized agents can take in a colony superstructure.

In addition to the development of specialized ant agents as described above, we explored two modes for general ant movement; {\it i)} ants were allowed to pick edges that could jump over layers in the colony (\ie, the superstructure is massively connected, with a plethora of skip connections), or {\it ii)} ants were only allowed to  select edges between consecutive layers (\ie, the superstructure is fully connected, with no skip connections). This was tested to see the impact that layer skipping would have on the sparsity and performance of generated RNNs.  
Jumping and non-jumping modes were tested for both the standard ants (with and without forward-path bias) and the specialized ant agent roles.
    



\subsection{Updating Pheromone Values}
\label{sec:pheromone_updates}
In this section, we describe the various schemes we experimented with in designing the ASNE optimization procedure. We define $\tau$ as the pheromone value, $\alpha$ as the pheromone decay parameter, $W$ as the weights of the evaluated (candidate) RNN, and $\eta$ as the candidate model's fitness. Specifically, we describe four different functional schemes used to model pheromone deposits.

The first strategy we implemented for ASNE is also standard for classical ACO setups. This deposit scheme rewards well performing RNNs with a fixed (constant) pheromone deposit while penalized ill-performing RNN models by evaporating the pheromone trace by a constant evaporation value, $C$. Specifically, this approach is defined as:
\begin{equation}\label{eq:fixed_constant}
	\tau_{new} =  \tau_{old} \pm C
\end{equation}
The second strategy we implemented was one that used the fitness (value) as a parameter to guide pheromone deposit. This has been shown to improve ACO performance in prior studies~\cite{sivagaminathan2007hybrid}. This scheme is defined as follows:
\begin{equation}\label{eq:reward_fit}
	\tau_{new} =  ( 1 - \alpha) \cdot \tau_{old} + \alpha \frac{1}{\eta}
\end{equation}
The third strategy was to use the values of the neural synaptic weights themselves to control/guide the deposit of pheromones. Specifically, we inserted a penalty on the weights, specifically an L1 penalty (assuming a Laplacian prior of the synaptic weight values), in order to encourage regularization that favored sparser connectivity structure. This form of weight decay is sometimes applied to ANNs when controlling for over-parameterization and sparse weight matrices (with many near hard-zero values) are highly desirable.
L1 regularization was applied to the pheromone deposition calculation in the following manner:
\begin{equation}\label{eq:reward_fit_weight_l1}
	\tau_{new} =  ( 1 - \alpha) \cdot \tau_{old} +
	\alpha \Big{\{}
									 \frac{1}{
									    \eta
									    +\frac{\gamma}{n}\|W\|
									}
			\Big{\}}
\end{equation}
The fourth and final strategy we employed was to insert an L2 penalty to regularize the RNN candidate weights. This assumes a Gaussian prior over the synaptic weight values and is sometimes referred to in ANN literature as ``weight decay''. We incorporate L2 regularization into pheromone deposition according to the following formula:
\begin{equation}\label{eq:reward_fit_weight_l2}
	\tau_{new} =  ( 1 - \alpha) \cdot \tau_{old} +
	\alpha \Big{\{}
									 \frac{1}{
									    \eta
									    +\frac{\gamma}{2n}\|W\|_2
									    }
			\Big{\}}
\end{equation}
We developed these L1 and L2 functional variations of pheromone deposit schemes in the hopes that they would ultimately encourage/reward the uncovery of sparse, compact RNN predictive models. 
    
\subsection{Pheromone Evaporation}
\label{sec:pheronome_evap}
Pheromone trace values (deposited on the superstructures synaptic edge pathways) evaporate or ``decay'' after each generation of an RNN in order to reduce the amount of pheromones on synaptic edges that are not being used much by ant agent collective~\cite{sivagaminathan2007hybrid,mavrovouniotis2013evolving,liu2006evolving}. Pheromone values are updated (or decayed) according to the following equation:
\begin{equation}\label{eq:evaporation}
    \tau_{updated} = (1 - \beta) \cdot \tau_{current} + \beta \cdot \tau_{original}
\end{equation}
where $\tau_{updated}$ is the pheromone value after the update, $\tau_{current}$ is the current pheromone value, $\tau_{original}$ is the original baseline pheromone value, and $\beta$ is the pheromone evaporation rate. This function evaporates the pheromone back towards the original baseline value.
        
\section{Results}
\label{sec:res}
    
All ASNE and \examm experiments generated $2000$ total RNNs, training each for $10$ epochs. NEAT, on the other hand, was allowed to generate $420,000$ RNNs. If we assume that a forward pass (forward propagation) and a backward pass (backprop calculation) are approximately the same computationally, this generously gave NEAT approximately $10$ times the amount of compute time (as $2000$ RNNs trained for $10$ epochs would equivocate to $20,000$ forward and $20,000$ backward passes). The RNNs with non-evolvable (fixed) architectures were allowed to train for $70$ epochs. Every experiment was repeated $10$ times to compute means and standard deviations in order to ensure a proper statistical comparison.

ASNE used a colony superstructure with $12$ input nodes, $3$ hidden layers, each with $12$ hidden nodes, and a single output node. 
Recurrent synapses could span $1$, $2$ or $3$ steps in time. The resulting connectivity superstructure consisted of $49$ nodes, $924$ edges, and $3626$ recurrent edges. While this may seem modest compared to modern convolutional architectures, which may consist of millions of connections, it is important to note that the RNNs generated from this superstructure are unrolled over $7200$ time steps (according to the time series length of the training and testing data samples) when trained locally via backpropagation through time (BPTT). This means algorithms such as ASNE must handle (fully-unrolled) networks of up to $3,528,000$ nodes, $6,652,800$ edges, and $26,107,200$ recurrent edges with errors from the final output (predictor) potentially back-propagated over up to $28,000$ synaptic connections.

The dataset utilized in this study is an open access time series dataset taken from a coal fired powerplant.  The data was introduced in previous neuro-evolution studies for time series data prediction~\cite{elsaid2019evolving,alex2019investigating}. 
It consists of $12$ possible parameters, recorded for $10$ days with each parameter recorded at each minute. These $12$ parameters were used to predict the flame intensity parameter (the response variable, in regression parlance). Results were generated by training RNNs on $5$ days worth of data taken from one of the coal burners from this data set. Fitness values (mean absolute error) were calculated on the other $5$ days, which was data that was treated as a test set.

$1,600$ experiments were conducted in order to include all combinations of the ASNE options/variations (described below). 
Each experiment was repeated $10$ times to obtain robust results. 
These ASNE experiments generated, trained, and evaluated $32$ million RNNs. Experiments were scheduled on a high performance computing cluster with $64$ Intel\textsuperscript{\textregistered} Xeon\textsuperscript{\textregistered} Gold 6150 CPUs, each with 36 cores and 375 GB RAM (total 2304 cores and 24 TB of RAM). Each experiment utilized $15$ nodes. Overall, it took approximately $30$ days to complete the entire battery of experiments.  Given the unstructured nature of the RNNs evolved in this work, utilizing CPUs has been found to be more efficient than GPUs as there are no wide, fully connected layers which would benefit from parallelized matrix algebra on a GPU.  Further, it allows the use of large scale high performance computing clusters which typically have many more CPUs than GPUs available.


\subsection{Backpropagation Hyperparameters}
\label{sec:bptt_params}
All ANNs were trained with backprop and stochastic gradient descent (SGD) using the same hyperparameters. SGD was run with a learning rate $\eta = 0.001$ and used Nesterov momentum ($mu = 0.9$) to smooth out the local gradient descent. 
No dropout regularization was used since it has been shown in other work to reduce performance when training RNNs for time series prediction~\cite{elsaid2018optimizing}. 
To prevent exploding gradients, gradients were re-scaled (as prescribed by Pascanu \etal~\cite{pascanu2013difficulty}) to a unit Gaussian ball when the norm of the gradient was above a threshold of $1.0$. To improve performance for vanishing gradients, gradient boosting (the opposite of clipping) was used when the norm of the gradient was below a threshold of $0.05$. 
The forget gate bias of the LSTM cells had $1.0$ added to it as this has been shown to yield significant improvements in training time by Jozefowicz \etal~\cite{jozefowicz2015empirical}. Weights for RNN in all other cases were initialized as described in the section describing our Lamarckian Weight Initialization~\ref{sec:lamarck} scheme for ASNE and in Ororbia~\etal~for EXAMM~\cite{ororbia2019examm}.

\subsection{ASNE Options and Hyper-parameters}
\label{sec:asne_parameters}

The influence/effect of individual ASNE hyper-parameters was carefully investigated in this study. A pheromone decay rate of $\alpha=0.05$ and a pheromone evaporation rate of $\beta = 0.1$ were chosen as they were shown to be effective in preliminary tests and is within the recommended standard range~\cite{sivagaminathan2007hybrid}. 
The other ASNE parameters we considered were:
    
\begin{enumerate}[i]
    \item Number of ants : \{20, 40, 80, 160\}.
    \item Regularization update parameter: \{0.25, 0.65, 0.90\}.
    \item Initializing RNN weights with constant $\Phi$ values of (\{0.3, 0.6, 0.9\}), using $\Phi$ as calculated by a function of fitness, and non-Lamarckian randomized weight initialization.
\end{enumerate}

\noindent
The application of the examined heuristics that appear in the figures and tables that follow are labeled as follows:

\begin{enumerate}[i]
    \item Function $\Phi$ : $\Phi()$
    \item Constant $\Phi$: $\Phi_{\text{\it value of }\Phi}$
    \item L1 Pheromone regularization: $L1_{\text{\it value of } \gamma}$ (Equation~\ref{eq:reward_fit_weight_l1})
    \item L2 Pheromone regularization: $L2_{\text{\it value of } \gamma}$ (Equation~\ref{eq:reward_fit_weight_l2})
    
    
    {
    \item Standard Ant Species:
        \begin{itemize}
            \item Standard Ants: $Std$
            \item Standard Ants with Bias: $StdBias$
        \end{itemize}
    }
    
    {
    \item Multi Species Ants:
        \begin{itemize}
            \item Explorer Ants: $Exp$
            \item Explorer Ants and Forward Social Ants: $ExpFwd$
            \item Explorer Ants and Backward Social Ants: $ExpBwd$
            \item Explorer Ants, Forward and Backward Social Ants: $ExpFwdBwd$
        \end{itemize}
    }
    \item Layer Jumping: $AJ$
    \item No Layer Jumping: $OJ$
    
\end{enumerate}
    



\subsection{Performance of Individual Heuristics}
\label{sec:indv_prefrom}


	
\begin{figure}[h!]
    \centering
    \includegraphics[angle=-90,origin=c, width=.5\textwidth]{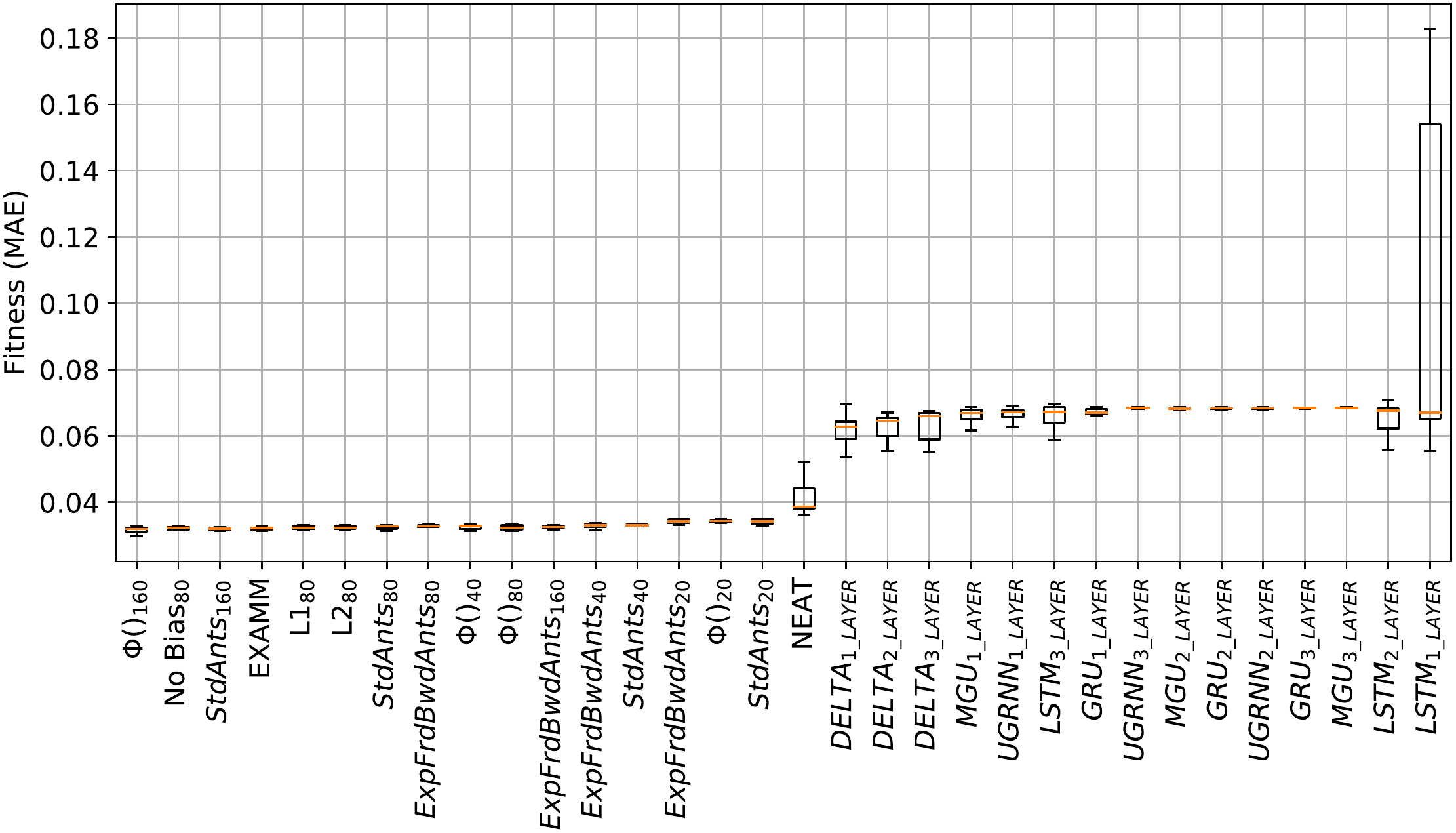}
	\caption{Performance of NEAT, EXAMM, \& individually applied ASNE heuristics against fixed memory cell RNNs.\label{fig:pure_unoptimized_fits}}
\end{figure}

Figure~\ref{fig:pure_unoptimized_fits} presents the performance of ASNE when each each heuristic is applied separately. Furthermore, it presents for comparison the performance of the state-of-the-art EXAMM, NEAT, and traditional fixed standard RNNs. While ASNE in this case (augmented by only one heuristic) did not outperform EXAMM except for some outliers, both EXAMM and ASNE showed dramatically better performance than NEAT, even though NEAT was given a significant amount of extra compute time. ASNE, EXAMM and NEAT also significantly outperformed traditional RNNs. Some of the gain over NEAT is most likely due to the use of backpropagation by EXAMM and ASNE since NEAT uses fairly simple and non-gradient based recombination operations to adjust weights.

\subsection{Performance of Combined Heuristics}
\label{sec:all_prefrom}
	
\begin{table*}[h]
\tiny
  \begin{center}
    \caption{Heuristic Ranking Statistics}
    \label{tab:heuristic_stats}
    \begin{tabular}{l|ccc|ccc|ccc|ccc|ccc}
    & \multicolumn{3}{c}{\textbf{Top 10}} & \multicolumn{3}{c}{\textbf{Top 25}}  & \multicolumn{3}{c}{\textbf{Top 100}}  & \multicolumn{3}{c}{\textbf{Top 250}} & \multicolumn{3}{c}{\textbf{Top 500}} \\
    & 
    Mean & Median & Best & 
    Mean & Median & Best & 
    Mean & Median & Best &
    Mean & Median & Best &
    Mean & Median & Best \\
    \toprule
    
    \textbf{$\Phi()$} 	& 3(0)	& 4(0)	& 3(0)	& 9(0)	& 7(0)	& 9(0)	& 26(0)	& 23(0)	& 31(8)	& 58(0)	& 54(0)	& 49(8)	& 108(1)	& 96(0)	& 100(14)		 \\
    \textbf{Const$\Phi$} 	& 7(0)	& 6(0)	& 7(0)	& 14(0)	& 14(0)	& 12(0)	& 60(0)	& 63(0)	& 54(8)	& 147(0)	& 149(0)	& 155(16)	& 294(0)	& 301(0)	& 299(43)		 \\
    \textbf{No$\Phi$} 	& 0(0)	& 0(0)	& 0(0)	& 2(0)	& 4(0)	& 4(0)	& 14(0)	& 14(0)	& 15(0)	& 45(0)	& 47(0)	& 46(0)	& 98(0)	& 103(0)	& 101(0)		 \\
    \textbf{L1} 	& 2(0)	& 4(0)	& 0(0)	& 9(0)	& 8(0)	& 3(3)	& 42(0)	& 34(0)	& 30(4)	& 96(0)	& 96(0)	& 91(4)	& 190(0)	& 186(1)	& 186(21)		 \\
    \textbf{L2} 	& 5(0)	& 5(0)	& 6(0)	& 13(0)	& 12(0)	& 16(1)	& 40(0)	& 45(0)	& 38(3)	& 100(0)	& 98(0)	& 95(12)	& 189(0)	& 192(0)	& 185(21)		 \\
    

    
     \textbf{StdAnts} 	& 0	& 0	& 0	& 1	& 0	& 0	& 3	& 0	& 0	& 20	& 19	& 0	& 80	& 77	& 7		 \\
    \textbf{StdBiasAnts} 	& 0	& 0	& 0	& 0	& 0	& 0	& 3	& 1	& 0	& 23	& 16	& 0	& 83	& 83	& 11		 \\

    \textbf{ExpAnts} 	& 0	& 0	& 10	& 0	& 0	& 25	& 1	& 0	& 100	& 10	& 6	& 250	& 92	& 85	& 440		 \\
    \textbf{ExpFrdAnts}	& 6	& 7	& 0	& 14	& 15	& 0	& 45	& 49	& 0	& 98	& 103	& 0	& 123	& 128	& 40 \\
    \textbf{ExpBkwAnts} 	& 0	& 0	& 0	& 0	& 0	& 0	& 0	& 0	& 0	& 0	& 0	& 0	& 0	& 0	& 0		 \\
    \textbf{ExpFrdBkwAnts} 	& 4	& 3	& 0	& 10	& 10	& 0	& 48	& 50	& 0	& 99	& 106	& 0	& 122	& 127	& 2 \\

    \textbf{No Jump} 	& 0	& 0	& 5	& 0	& 0	& 13	& 0	& 0	& 52	& 0	& 0	& 128	& 2	& 9	& 282		 \\
    \textbf{Layer Jump} 	& 10	& 10	& 5	& 25	& 25	& 12	& 100	& 100	& 48	& 250	& 250	& 122	& 498	& 491	& 218	\\
    \textbf{20 Ants} 	& 0	& 0	& 2	& 0	& 0	& 6	& 0	& 0	& 24	& 0	& 0	& 65	& 3	& 6	& 220		 \\
    \textbf{40 Ants} 	& 2	& 0	& 3	& 5	& 1	& 7	& 14	& 15	& 23	& 50	& 57	& 63	& 97	& 87	& 120		 \\
    \textbf{80 Ants} 	& 4	& 3	& 2	& 8	& 11	& 6	& 44	& 45	& 26	& 82	& 80	& 60	& 175	& 173	& 80		 \\
    \textbf{160 Ants} 	& 4	& 7	& 3	& 12	& 13	& 6	& 42	& 40	& 27	& 118	& 113	& 62	& 225	& 234	& 80		 \\

\end{tabular}
\end{center}
\end{table*}	

\begin{figure}[h!]
    \includegraphics[angle=-90,origin=c, width=.98\textwidth]{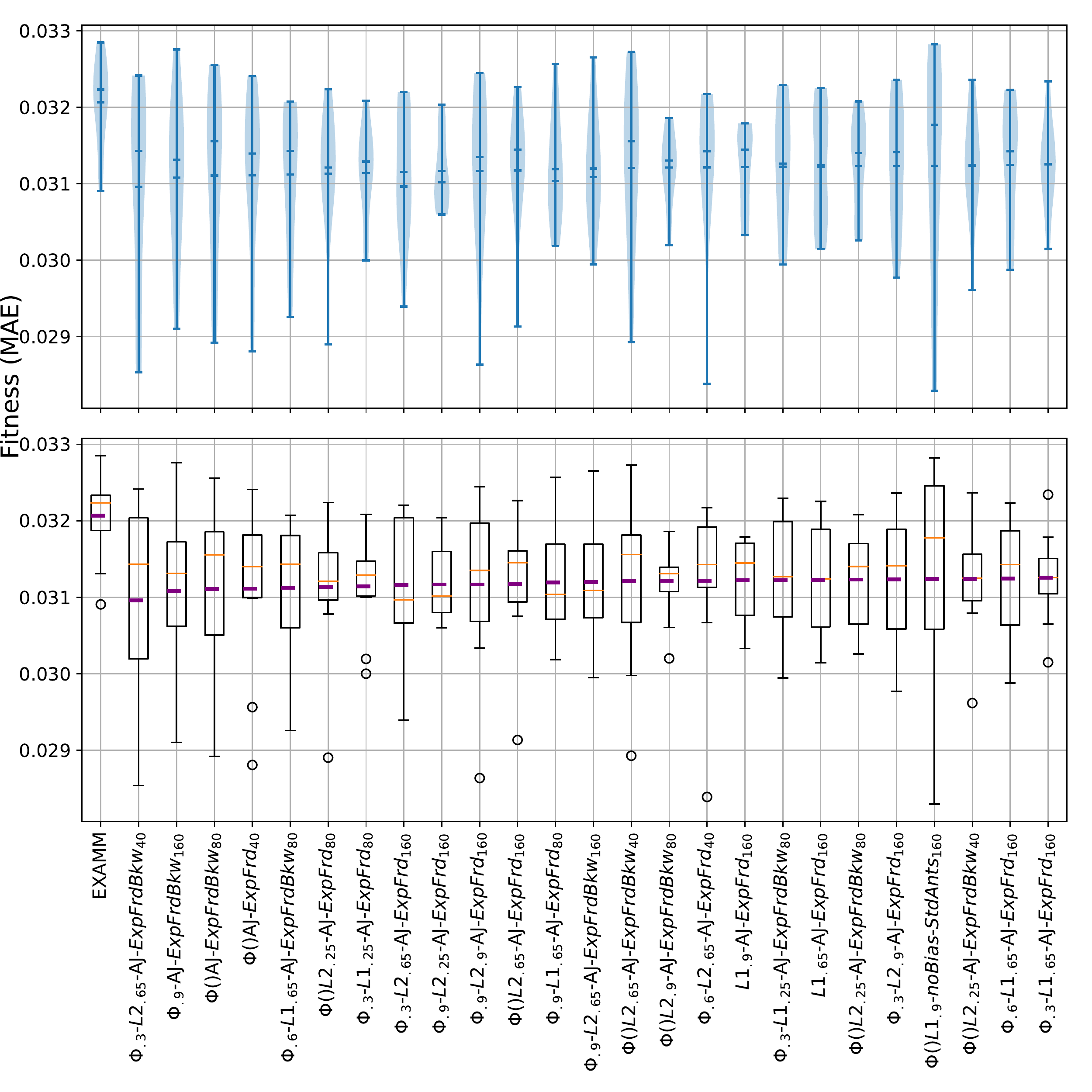}
    \caption{Performance of EXAMM and the top 25 ASNE experiments \label{fig:bp_vio_fit}}
\end{figure}

The combined application of multiple different heuristics, as illustrated in Figure~\ref{fig:bp_vio_fit}, yielded ASNE results that outperformed \emph{all baselines}, including the fixed RNNs, NEAT, \emph{as well as} EXAMM. 
Table~\ref{tab:heuristic_stats} provides statistics ranking each of the heuristics based on how many times the experiments that utilized them appeared in the top $10$, $25$, $100$, $250$, and $500$ best results as determined by the mean, median, and the best performance of the RNN generated in the experiment's 10 repeats. Values in parentheses are the number of times an experiment that only utilized that heuristic appeared in that top ranking. 
The utilization of multiple heuristics dominated the top results, with individually-applied heuristics not appearing in the top $10$, and only $4$ times in the top $25$ (only as best results).

Lamarckian weight inheritance also proved to be important, yielding strong performance, with all of the top $10$ utilizing either functional or constant $\Phi$ parameters. Furthermore, it also occurred $2$ (mean), $4$ (median) and $4$ (best) times in the top $10$, and $14$ (mean), $14$ (median), and $15$ (best) times in the top $25$.

Additionally, all of the best performing RNNs used layer-jumping ants, which tend favor more sparse connectivity patterns. Most of the best results used pheromone weight-regularization, with L2 regularization appearing at a nearly 50\% rate in the top $10$, $25$ and $100$ results. The regularization factor was also high, at 65\% or 90\%, for most of the $25$ best experiments that used it.

All of the top $250$ best results utilized the multiple ant species heuristic, which strongly supports the use of specialized ants. The number of ants varied between $20$ and $160$ for all the top $25$ results in the mean and median case, with a larger number of ants tending to perform better. However the case of $20$ ants did occasionally appear in the best cases, even sometimes in the top $10$ and, furthermore, these networks tended to be rather sparse but very well performing. This may suggest that the experiments that utilized more ants had an easier time finding the most important structures, but also potentially had extraneous connections which were not needed. In contrast, the experiments with less ants had less of a chance of finding these important structures due to lower (overall) connectivity. This suggests that further optimizations could be designed to better guide ASNE towards the discovery of more efficient network architectures.

Perhaps one the most interesting items to observe is the performance distribution when multiple ant agent roles was used in ASNE. The entirety of the best found RNNs, up to the top $250$ were from explorer ants only, so these generated RNNs only had recurrent connectivity in terms of whatever the various memory cells offered. However, for the mean and median performance of the experiments, nearly all the top $25$, $100$, and $250$ consisted of explorer and forward recurrent roles or explorer, forward, and backward recurrent ant specializations -- with only a very few of the only explorer ant only configurations showing up in the top $100$ and $250$. First, this suggests that backward recurrent connections (which are most commonly utilized in RNNs) were less effective than forward recurrent connections. Second, it also appears that adding these recurrent connections tended to make the RNNs perform significantly better on the average and median cases, while the RNNs which were generated with only explorer ants had the ability to occasionally find RNNs that generalized quite well.  These results certainly suggest further study in order to better understand the effect of combining recurrent connections and memory cells. In addition, perhaps alternative strategies can be developed that retain the stability of adding recurrent connections while still efficiently finding well-generalizing RNNs.
	
\paragraph{RNN Density}
\label{sec:density}
Tables~\ref{tab:no_edges} and \ref{tab:no_rec_edges} show the number of nodes, edges, and recurrent edges in the best evolved RNNs for the experiments related to ASNE augmented only with single, individual heuristics. EXAMM found the simplest structures but these were not always the best-performing, which may suggest that EXAMM, as powerful as it is, still sometimes gets trapped in local minima. Utilizing the multiple ant agent roles and L2 pheromone regularization proved to be very effective in generating smaller, sparser RNNs. The smaller RNN size combined with its strong performance in the top rankings suggest that modeling ant role specialization can significantly improve how well an ACO/neuro-evolutionary, such as the proposed ASNE, generates candidate RNNs.

\begin{figure}[h]
    \centering
    \includegraphics[angle=-90,origin=c,width=.6\textwidth, height=.55\textheight]{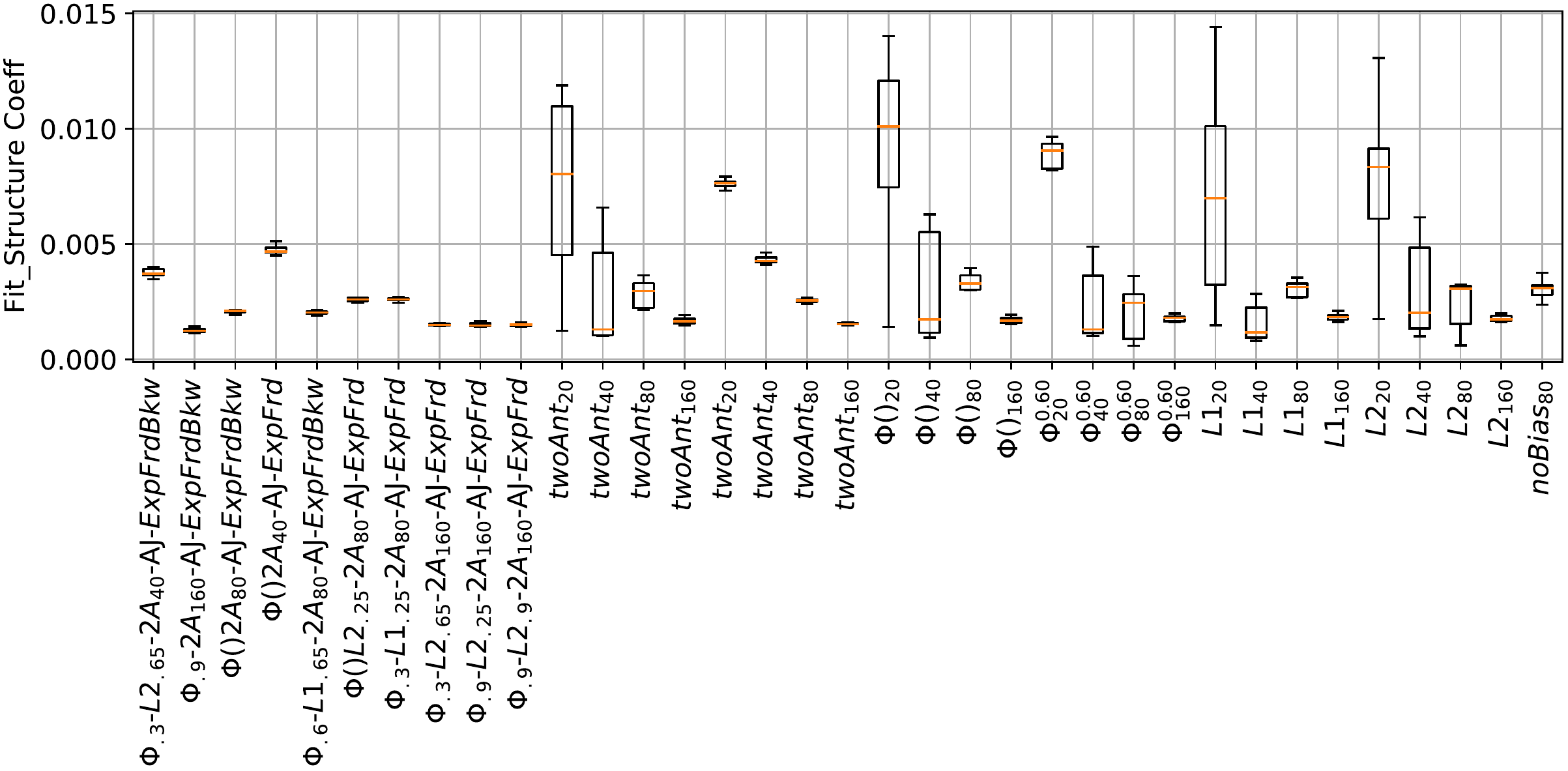}
    \caption{Ranges of how much each weight in an RNN contributed to its performance (Equation~\ref{eq:fit_strct_coeff}). }
    \label{fig:fit_strct_coeff}
\end{figure}

\paragraph{Fitness Structure Coefficient}
Figure~\ref{fig:fit_strct_coeff} examines the relationship between the size of the network and its fitness. Results for RNNs from the top $10$ best performing experiments are shown along with RNNs (taken from the individual heuristic experiments).

The following equation was used to calculate a measure of the contribution of each weight to the fitness of the RNN:
\begin{equation}
    \label{eq:fit_strct_coeff}
    C_{fit\_struct} = \frac{1-MAE}{W_S}
\end{equation}
where $C_{fit}$ is a structural coefficient calculated by $MAE$ (the mean absolute error of the RNN) and $W_S$ is the number of weights currently contained in the candidate RNN structure. Higher values represent RNNs where weights contribute more to the performance of the network. 

	\begin{table}[!h]
	  \begin{center}
	    \caption{Number of Edges (40 Ants)}
	    \label{tab:no_edges}
	    \begin{tabular}{lccccc} 
	      \textbf{} & \textbf{Min} & \textbf{Max} & \textbf{Avg} & \textbf{$\sigma$} & \textbf{Reduce\%} \\
	      \toprule
	      
	        $\Phi$	                & $54$	& $134$	    & $85.7$	& $26.77$	& $90.73$\%	 \\
            $Std$	            & $61$	& $144$ 	& $109.8$	& $29.89$	& $88.12$\%	 \\
            $StdBias$   	    & $54$	& $133$	    & $100.8$	& $27.35$	& $89.09$\%	 \\
            $Exp$	            & $109$	& $122$	    & $116.1$	& $4.01$	& $87.44$\%	 \\
            $ExpFrd$	        & $106$	& $122$	    & $113.9$	& $5.54$	& $87.67$\%	 \\
            $ExpBwd$	        & $105$	& $118$	    & $112.1$	& $4.32$	& $87.87$\%	 \\
            $ExpFrdBwd$	    & $106$	& $117$	    & $113.6$	& $3.69$	& $87.71$\%	 \\
            L1	                    & $77$  & $142$	    & $112.8$	& $21.36$	& $87.79$\%	 \\
            L2	                    & $61$  & $156$	    & $102.1$	& $35.16$	& $88.95$\%	 \\
            \examm	                & $26$	& $113$	    & $52.2$	& $25.65$	& $94.35$\%	 \\
            
	        
	    \end{tabular}
	    \vspace{-0.4cm}
	  \end{center}
	\end{table}
	
	\begin{table}[h!]
	  \begin{center}
	    \caption{Number of Recurrent Edges (40 Ants)            }
	    \label{tab:no_rec_edges}
	    \begin{tabular}{lccccc} 
	      \textbf{} & \textbf{Min} & \textbf{Max} & \textbf{Avg} & \textbf{$\sigma$} & \textbf{Reduce \%} \\
	      \toprule
	        $\Phi$	                & $76$	& $879$	& $348.2$	& $281.12$	& $90.40$\%	 \\
            $Std$	& $87$	& $873$	& $575.0$	& $315.44$	& $84.14$\%	 \\
            $StdBias$	    & $85$	& $818$	& $508.1$	& $308.38$	& $85.99$\%	 \\
            $Exp$	        & $0$	& $0$	& $0.0$	    & $0.00$	&   -   	 \\
            $ExpFrd$	        & $62$	& $73$	& $69.3$	& $2.90$	& $98.09$\%	 \\
            $ExpBwd$	        & $47$	& $54$	& $50.7$	& $2.33$	& $98.60$\%	 \\
            $ExpFrdBwd$	    & $97$	& $118$	& $110.1$	& $5.49$	& $96.96$\%	 \\
            L1	                    & $93$	& $849$	& $577.5$	& $222.58$	& $84.07$\%	 \\
            L2	                    & $86$	& $972$	& $518.5$	& $364.21$	& $85.70$\%	 \\
            \examm	                & $1$	& $14$	& $9.6$	    & $3.47$	& $99.74$\%	 \\

	      
	    \end{tabular}
	    \vspace{-0.5cm}
	  \end{center}
	\end{table}

\section{Discussion}
\label{discuss}

To the best of our knowledge, this work represents the first application of ant colony optimization (ACO) to the problem of neuro-evolution/neural architecture search for recurrent neural networks with varying recurrent time spans and more complex connectivity patterns (the only prior related study that investigated ACO for evolving RNNs was critically constrained to small RNNs with a single recurrent timestep and Elman-style connectons~\cite{desell2015evolving}). Specifically, we proposed the novel ant swarm neuro-evolution algorithm (ASNE) for metaheuristically searching the the massive search space of possible RNNs with complex connectivity patterns (of both recurrent and feedforward forms). ASNE generates candidates from a massively-connected superstructure (the colony/swarm), taking advantage of ACO for structural optimization and concepts from neuro-evolutionary/genetic approaches for maintaining populations of RNN candidates that are trained locally and asynchronously (making ASNE a memetic procedure as well). A hallmarkk of ASNE is its computational formalization of role/specialization in real ant colonies --  ant agents internally are prevented from getting stuck ``wandering'' around the superstructure through the use of different kinds of ant agents that are constrained to only explore different components of the underlying complex graph space. This is a form of modularization that proves particularly useful in cutting up large, complex search spaces under ASNE.

Our experimental results show that using ants with different roles generated RNNs that were not only sparse but performant -- these candidates almost entirely outperformed the more standard ant traversal strateg even when standard ants were biased to more likely select forward paths. This innovation of utilizing multiple ant types improves the ACO core of ASNE when searching for effective RNNs. Furthermore, Lamarckian weight inheritance greatly improved the accuracy of the generated RNNs\footnote{Corraborating prior studies that have also shown the benefits of such an initialization scheme~\cite{desell2018accelerating,ororbia2019examm}.} and allowing ants to jump (or skip) layers proved to not only boost performance but also to increase sparsity.
Lastly, to our knowledge the introduction of L1 and L2 regularization into the ACO pheromone deposition process is quite novel if albeit a bit unconventional. Our results show by playing with the form of the pheromone adjustment function, we can increase the likelihood that sparser RNNs are found that also outperform schemes that do not incorporate regularization/constraints. The strategies we formalize in this work are generic and could be applied to any other ACO algorithm's pheromone update process.

The proposed ASNE metaheuristic not only provides advances and new concepts for the field of ant colony optimization research to further explore but also shows strong promise for its use as an alternative neuro-evolution algorithm for automated RNN architecture search. It significantly outperforms the well-known NEAT algorithm (even when NEAT is given an order of magnitude more computation), and, more importantly, ASNE outperforms the state-of-the-art EXAMM genetic evolutionary algorithm on the time series problem studied in this paper. 

The work also opens up a number of avenues for future study as well as presents some interesting questions.  In particular, why were explorer ants able to find the best networks and yet performed quite poorly in the mean and median cases? Why did explorer ants combined with social recurrent ants perform extremely well in the mean and median cases but not in the best cases? Answering experimental queestions such as these might lead to insights as to how recurrent connections that skip multiple steps of time interact with recurrent memory cells, potentially leading to the design of more expressive, RNN structures that better capture longer-term dependencies in sequential data. Finally, future work should entail investigation of ASNE on other time series datasets as well as sequence modeling (and classification) problems more commonly explored in mainstream statistical learning research, such as language modeling \cite{mikolov2010recurrent,ororbia2017learning}.

\section*{Acknowledgements}
This material is in part supported by the U.S. Department of Energy, Office of Science, Office of Advanced Combustion Systems under Award Number \#FE0031547. We also thank Microbeam Technologies, Inc. for their help in collecting and preparing the coal-fired power plant dataset.
Most of the computation of this research was done on the high performance computing clusters of Research Computing at Rochester Institute of Technology. We would like to thank the Research Computing team for their assistance and the support they generously offered to ensure that the heavy computation this study required was available.

\bibliographystyle{unsrt} 
\bibliography{bibliography.bib} 

\newpage
\begin{appendices}
\section{Complete Results}
\begin{longtable}{lccccc} 
    \textbf{} & \textbf{Minimum} & \textbf{Maximum} & \textbf{Mean} & \textbf{Median} & \textbf{Std. Dev.} \\

NEAT 	& $0.0364212$ 	& $0.0531144$ 	& $0.04548$ 	& $0.04838$ 	& $0.00631$ 	\\
EXAMM 	& $0.0309058$ 	& $0.0328495$ 	& $0.03207$ 	& $0.03223$ 	& $0.00056$ 	\\
$\Phi.3$-L2.65-$2Ant_{40}$-AJ-ExpFrdBkw 	& $0.028535$ 	& $0.0324159$ 	& $0.03096$ 	& $0.03143$ 	& $0.00137$ 	\\
$\Phi.9$-$2Ant_{160}$-AJ-ExpFrdBkw 	& $0.0291028$ 	& $0.0327583$ 	& $0.03108$ 	& $0.03131$ 	& $0.00102$ 	\\
$\Phi()$-$2Ant_{80}$-AJ-ExpFrdBkw 	& $0.0289199$ 	& $0.0325558$ 	& $0.03111$ 	& $0.03155$ 	& $0.00113$ 	\\
$\Phi()$-$2Ant_{40}$-AJ-ExpFrd 	& $0.028806$ 	& $0.0324082$ 	& $0.03111$ 	& $0.03140$ 	& $0.00106$ 	\\
$\Phi.6$-L1.65-$2Ant_{80}$-AJ-ExpFrdBkw 	& $0.0292576$ 	& $0.0320735$ 	& $0.03112$ 	& $0.03143$ 	& $0.00090$ 	\\
\end{longtable}
\end{appendices}

\end{document}